\documentclass{article}
\pdfoutput=1
\usepackage{amsmath}
\usepackage{amssymb}
\usepackage{epsfig}
\usepackage{subfig}
\usepackage{graphics}
\usepackage{wrapfig}
\usepackage{nips14submit_e,times}
\usepackage{hyperref}
\usepackage{url}
\usepackage{bbm,appendix}
\usepackage{color}
\title{\textcolor{black}{Locally} Scale-Invariant Convolutional Neural Networks}
\author{
Angjoo Kanazawa\\
Department of Computer Science\\
University of Maryland,\\
College Park, MD 20740 \\
\texttt{kanazawa@umiacs.umd.edu} \\
\And
Abhishek Sharma\\
Department of Computer Science\\
University of Maryland\\
College Park, MD 20740 \\
\texttt{bohkaal@umiacs.umd.edu} \\
\AND
David Jacobs\\
\texttt{djacobs@umiacs.umd.edu} \\
Department of Computer Science\\
University of Maryland\\
College Park, MD 20740 \\
}

\nipsfinalcopy

\begin{document}
\maketitle
\begin{abstract}
\textcolor{black}{Convolutional Neural Networks (ConvNets) have shown excellent results on many visual classification tasks. With the exception of ImageNet, these datasets are carefully crafted such that objects are well-aligned at similar scales. Naturally, the feature learning problem gets more challenging as the amount of variation in the data increases, as the models have to learn to be invariant to certain changes in appearance. Recent results on the ImageNet dataset show that given enough data, ConvNets can learn such invariances producing very discriminative features \cite{alex}. But could we do more: use less parameters, less data, learn more discriminative features, if certain invariances were built into the learning process? In this paper we present a simple model that allows ConvNets to learn features in a locally scale-invariant manner \emph{without} increasing the number of model parameters. We show on a modified MNIST dataset that when faced with scale variation, building in scale-invariance allows ConvNets to learn more discriminative features with reduced chances of over-fitting.} \end{abstract}

\section{Introduction}
Convolutional Neural Networks (ConvNets) \cite{lenet} have achieved excellent
results on visual classification tasks like handwritten digits
\cite{MNIST}, toys \cite{norb}, traffic signs \cite{traffic}, and recently
1000-category ImageNet classification \cite{alex}.
ConvNets' success comes from their ability to learn complex patterns
by building increasingly abstract representations layer by layer, much like other deep neural networks.
However, ConvNets differ in that they exploit the two dimensional
structure of images where objects and patterns appear at arbitrary
locations. ConvNets apply local filters at every position in the image,
allowing the network to detect and learn patterns regardless of their location.

In reality, the world has a three dimensional structure, and objects
at different distances will appear in an image at different scales as
well as locations. ConvNets do not have a mechanism to take advantage of scale explicitly,
so to detect a single pattern at multiple scales, they must learn to separate filters for each scale.
Unfortunately this has several major shortcomings. Capturing
patterns at multiple scales uses up resources that could be used to
learn a wider variety of feature detectors. This requires an increase in the
number of feature detectors, which means that the network is
harder to train, takes longer to train, and is more likely to overfit.
To learn multiple scales and prevent overfitting, you would need a
lot of training data, and even then the network will only respond to
the scales seen during training. 
Also, detectors that capture the same pattern but at different scales
are learned independently without sharing the training
samples. Finally, having multiple filters of a single pattern at
different scales burdens the next layer by increasing the number of
configurations that indicate the presence or absence of that pattern.

In this paper, we present \emph{scale-invariant} convolutional networks (SI-ConvNets),
which applies filters at multiple scales in each layer so a single filter
can detect and learn patterns at multiple scales. We max-pool
responses over scales to obtain representations that are
locally scale invariant yet have the same dimensionality as a traditional
ConvNet layer output. The proposed architecture differs from other multi-scale approaches
explored with ConvNets since scale-invariance is built in at the layer
level rather than at the network level. We also achieve locally
scale-invariant representation and  we do not
require any increase in the number of parameters to be learned. We show in our experiments that by sharing
information from multiple scales, the proposed model can achieve
better classification performance than ConvNets while simultaneously requiring less training data.
We evaluate the proposed model on a variation of the MNIST dataset where digits appear at multiple scales,
and demonstrate that the SI-ConvNets are more robust to scale
variations in training data and unfamiliar scales in test data than ConvNets.
Our model is complementary to other ConvNet architectures and it can easily be incorporated into existing variants,
and we will make the source code available online for the research community.

\section{Background}
\label{cnn}
Several recent works have addressed the problem of explicitly incorporating transformation invariance in deep learning models. In the unsupervised feature
learning domain, Sohn \emph{et al.} \cite{honglak} and \cite{kivinen} introduced
transformation-invariant Restricted Boltzmann Machines (RBMs) where linear transformations of a
filter are applied to each input to infer its highest activation. In these two models, the transformed filters were only applied at the center of the
largest receptive field size. Our model uses an inverse transformation stage so that transformed filters can be applied densely but still retain correspondence.
Our work is inspired by the success of Sohn \emph{et al.} and our goal
is to incorporate scale-invariant feature learning into the extremely
successful ConvNet models \cite{alex}.

Tiled convolutional neural networks \cite{tiled} learn invariances implicitly 
by square-root pooling hidden units that are computed by
partially un-tied weights.  In comparison, our approach explicitly encodes scale invariance,
and does not require the increase in the number of learned parameter that
is required by un-tying of weights.
\cite{yanworkshop} fuses outputs of multiple ConvNets applied over
multiple scales for semantic segmentation, but each ConvNet is learned independently without weight-sharing. In
contrast, we jointly learn a set of feature detectors that are shared
over multiple scales. \cite{traffic} proposes a multi-scale ConvNet where outputs of all
convolutional layers are fed to the classifier. This enables them to
capture information from different levels of the hierarchy, but there is no
scale invariance in the features learned at a layer because each layer is only
applied to the original scale.

Another influential work is that of Farabet \emph{et al.}\cite{farabet}, who train
ConvNets over the Laplacian pyramid of images with tied weights for scene parsing problems. In their model, the
entire multi-layer forward propagation is applied end to end at each scale
disjointly, then right before the fully-connected layers, the responses of all scales are aligned by up-sampling and concatenated. Keeping responses from each scale allows them to capture scale-level dependencies, but at the expense of increasing the number
of parameters required in the final layer. This further restricts the number of
scales that can be applied; in contrast, our model is more compact and
free of such restrictions. 
Further, their approach is motivated by the need to have large receptive field sizes to capture long-range, contextual interactions
that are necessary for scene understanding. 
In contrast, we are interested in capturing locally scale-invariant features that are useful for
image classification; hence, unlike their approach, we pool the
responses over all scales in each spatial location in each layer. \textcolor{black}{Pooling responses over scales in each layer as opposed to concatenating all scales in the very end has subtle but different effects in the middle layers. For example, suppose there is an image of two circles of different sizes, and a circle filter that can detect one of the circles in the image. In the architecture of Farabet \emph{et al.}, each circle will be detected in a different scale, but they will never be recognized together until the layer where all scales are concatenated. In our architecture, both circles can be detected together and the second layer can immediately make use of the fact that there are two circles in the image. Of course, by having two circle filters of different sizes, Farabet \emph{et al.}'s network can also detect two circles in one scale, but at the expense of learning redundant filters.} 
While their work demonstrates the advantages of applying ConvNets over multiple scales in scene parsing, we further investigate its effectiveness in an image classification domain with a more modular model that explicitly incorporates scale-invariance in each layer.

\subsection{Convolutional Neural Networks}
\label{sec:conv-neur-nets}
Convolutional Neural Networks (ConvNets) are a supervised feed-forward
multi-layer architecture where each layer learns feature detectors of
increasing complexity. The final layer is a classifier or regressor
with a cost function such that the network can be trained in a
supervised manner. 
The entire network is optimized jointly via stochastic gradient descent with gradients computed
by back-propagation \cite{lenet}. A single layer in a ConvNet is
usually composed of feature extraction and nonlinear activation stages,
optionally followed by spatial pooling and feature normalization.

The hallmark of ConvNets is the idea of convolving local feature
detectors across the image. This idea is motivated by the fact that similar patterns can appear anywhere in the 2D layout of
pixels and that nearby values present strong dependencies in
natural images \cite{yoshua}. The local feature detectors are
trainable filters (kernels) and their spatial extent is called the \emph{receptive
  field} of the network, as it represents how much of the image the network gets to
``see''. The convolution operation effectively ties the learned weights at multiple locations, which 
radically reduces the number of trainable parameters as compared to having different weights
at each location \cite{lenet}. The output of convolving one kernel is called
a \emph{feature map}, which is sent through a nonlinear activation
function $\sigma$.  A feature map $h$ at a layer is computed as
\begin{equation}
  h = \sigma( (W \ast x) + b),
\end{equation}
where $\ast$ is the convolution operator, $x$ is the input feature map from the previous layer, $W$ and $b \in \mathbb{R}$ are the trainable weights and bias
respectively. By having multiple feature
maps, the network can represent multiple concepts in a single
layer. The model may further summarize each sub-region of the feature map via max or average
pooling, providing invariance to small amounts of image translation.

\section{Scale-Invariant Convolutional Neural Network}
\label{sec:sicnn}
\begin{figure*}[tbp]
\centerline{\resizebox{0.8\textwidth}{!}{\includegraphics {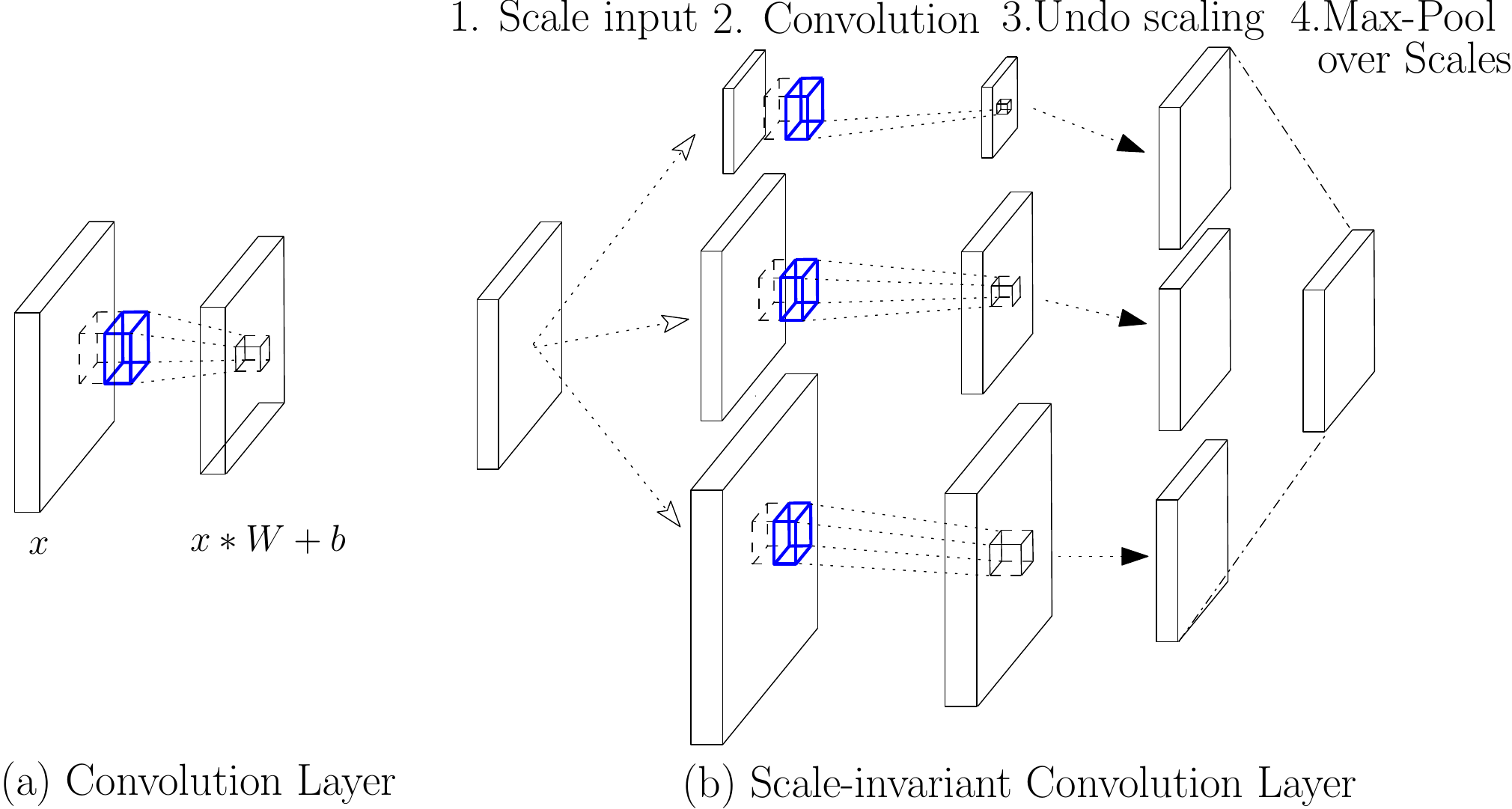}}}
\caption{\small{Side-by-side comparison of the structure of a (a) convolution
  layer and the proposed (b) \emph{scale-invariant} convolution
  layer. The black bold box is the
  feature detector kernel in both figures.  In (b), first, the image pyramid is
  constructed by scaling the input (only 3 scales shown for brevity).
  Second, the same kernel is convolved across the input in each scale. Third, the responses of the
convolution in each scale are normalized. Finally,  responses are pooled over scale at each spatial
location obtaining a locally scale-invariant representation. Note the
input and output size of (a) and (b) are the same. }}
\label{fig:arch}
\end{figure*}

Feature detectors in ConvNets have the ability to
detect features regardless of their spatial locations in the image,
but the same cannot be said for features at different
scales. 
In this section we describe a scale-invariant ConvNet
(SI-ConvNet).  Our formulation also allows the output of ConvNets to be
locally scale-invariant, where the representation of the same patterns
at different scales will be similar \footnote{Given that the patterns
  share the same center. When the center of the patterns are shifted, the output
  will be similar but at different locations, i.e. a shift-equivariant
representation}. 
Figure \ref{fig:arch} shows the side by side comparison of the overall structure of these two layers.

\subsection{Forward Propagation}
\label{sec:fprop}
Our goal is to let one feature detector respond to patterns at
multiple scales. To do so, we convolve the filters over multiple resolutions
in a pyramid. At each scale the exact same filters are used to convolve
the image (weight-tying). Since we spatially transform the image, the
outputs of convolution come in different spatial sizes. In order to
align the feature maps, we apply an inverse
transformation to each feature map \footnote{When the stride
is equal to the size of the kernel, applying the inverse transformation
gives direct correspondence between convolution outputs. When it's not,
after applying the inverse transformation, the output has to be
either cropped or padded with 0s to be properly aligned.}. Finally we max-pool
the responses over all the scales at each spatial location. Pooling
responses over multiple scales serves two purposes. First, it
 allows us to obtain a locally scale-invariant representation. Second, it
summarizes the responses in a concise way that allows us to maintain
the same output size as a standard convolution layer.

 Specifically, let $\mathcal{T}$ be a linear image transform operator that applies some spatial transformations to an input $x$. Then, for a set of $n$ transformation operators $\{\mathcal{T}_1, \dots, \mathcal{T}_n\}$,
a feature map, $h$, is computed as:
\begin{align}
  \label{eq:sicnn}
 \hat z_i &= (W \ast \mathcal{T}_i(x)) + b\\
 z_i &= \mathcal{T}_i^{-1}(\hat z_i)\\
 h &= \sigma\left (\max_{i\in \{1, \dots, n\}} \, [z_i]\right ).
\end{align}
Note that $\hat z_i \in \mathbb{R}^{h_i \times h_i \times m}$, where $h_i$ is
the size of the output of convolution in the $i$-th
transformed input, and $z_i \in \mathbb{R}^{h_0 \times h_0 \times m}$
for all $i$, where $h_0$ is the canonical output size when all
responses are aligned and $m$ is the number of feature maps
used. $\mathcal{T}_1$ is always the identity transformation, so when
$n=1$ the framework is equivalent to traditional ConvNets.

\begin{figure*}
\centerline{\resizebox{\textwidth}{!}{\includegraphics {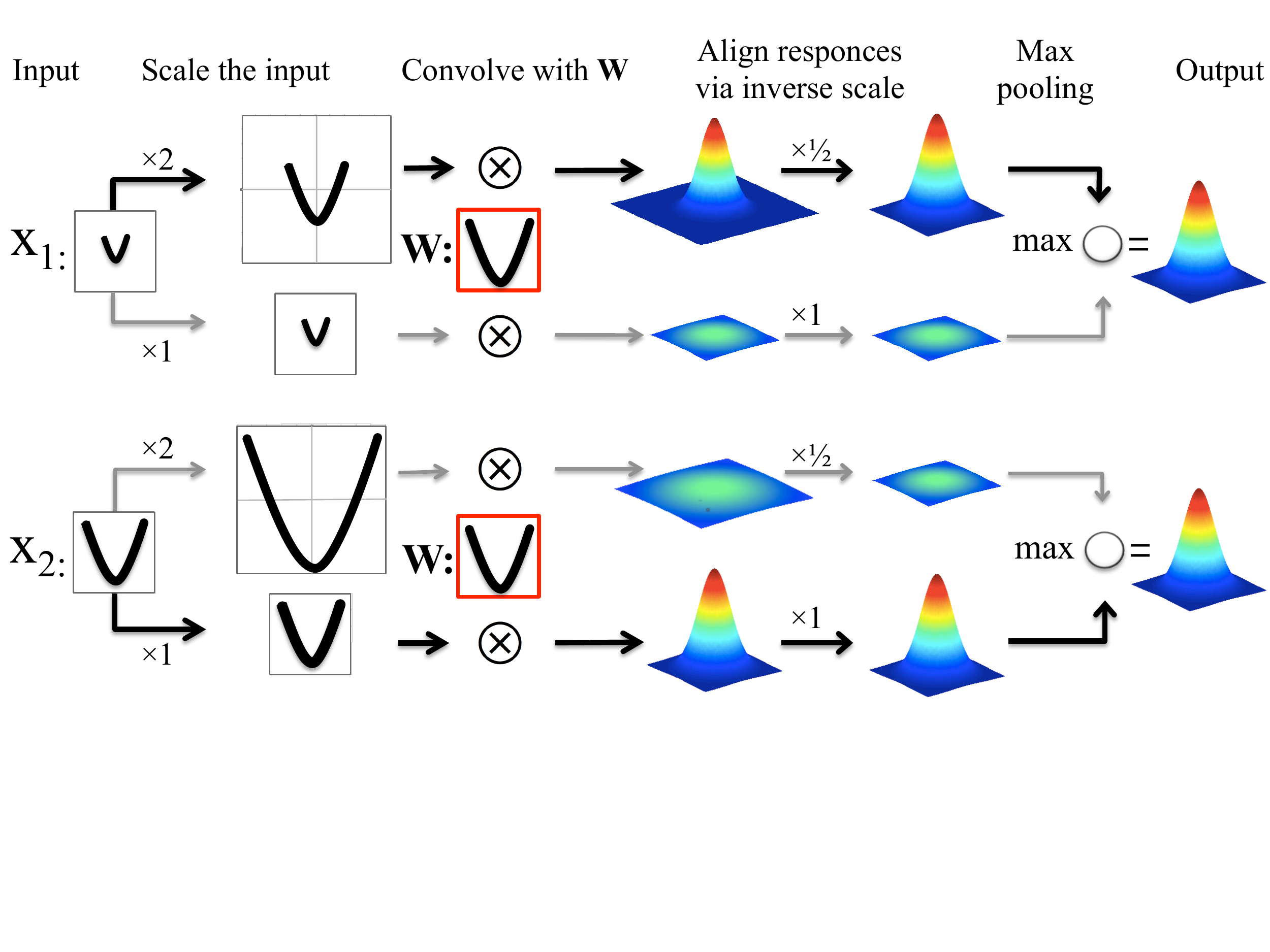}}}
\caption{An illustration showing that a single feature detector $W$ can detect the
  features in images $x_1$ and $x_2$ despite their scale difference: In a standard convolution layer, $W$ will only highly
respond to $x_2$ because of the scale difference between $W$ and $x_1$
. However, in a scale-invariant convolution layer, $W$ can respond
well to $x_1$ when it is scaled to twice its original size. Thus both
$x_1$ and $x_2$ achieve similar feature representation. The bold arrows indicate the winning scale that is used in the final
representation. Only two scales are shown here for illustration purposes.}
\label{fig:example}
\end{figure*}

Figure \ref{fig:example} illustrates this idea with sample inputs $x_1$
and $x_2$ that have the same ``V'' pattern but at different
scales. $W$ is a ``V'' detector that this ConvNet has learned. With this $W$, a standard
convolution layer will only activate $x_2$ whose ``V'' pattern matches
the size of ``V'' in $W$. However, in a scale-invariant convolution layer, $x_1$ and $x_2$
undergo scale transformations where $W$ can be matched in one of the
scales, allowing $W$ to detect the pattern on both $x_2$ and $x_1$. The responses are aligned via inverse
transformation and the final output is the maximum activation at each
spatial location. The path of the winning scale is shown in bold lines.

Since convolving a pyramid of images with a single filter is analogous
to convolving a single image with filters of different
sizes, using $n$ scales is analogous to increasing the number of feature
maps by $n$ times without actually paying the price of having
more parameters. This allows us to train a more expressive model without increasing the chances for over-fitting.

The image transform operator $\mathcal{T}(x)$ is parametrized by a
single scale factor $s$ where each 2D location $\vec x$ in the
transformed image is computed by a linear interpolation of the original image around $s^{-1}\vec{x}$.
We use bilinear interpolation to compute the coefficients. Note that while we focus on scale invariance in this
paper, the framework is applicable to other linear transformations. The transformation coefficients can be precomputed, so applying the transformation is efficient. However, there is an increase in the number of convolution operations required since each scaled
input must be convolved. The increase is dominated by the largest
scale factor $s_n$ and the step sizes $r > 1$ between each
scale. Explicitly, the increase in the number of convolutions in a layer is $s_n^2[\frac{1-(1/r)^n}{1-(1/r)}]$. Please refer to the supplementary
material for details.

\subsection{Backward Propagation}
\label{sec:backprop}
Since the scale-invariant convolution layer consists of just linear and max
operations, its gradients can be computed by a simple modification of the back-propagation algorithm. Backprop for max-pooling over scale is implemented by using the argmax indices, analogous to how backprop is done for spatial max-pooling. For scale transformations, the error signal is propagated through the bilinear coefficients used to compute the transformation. Please refer to the supplementary materials for the detailed derivations.

\section{Experiments}
\label{experiments}
\vspace{-0.2cm}
We first compare the performance of the proposed method, referred to as
``SI-ConvNet'', against other baseline methods including traditional ConvNets. For all the experiments we carried out, all networks share the exact same hyper-parameters and architecture, except that the convolution layers that are replaced by scale-invariant convolution layers.  We implement our method using the open-source Caffe framework \cite{caffe}, and our code will be available online.

\textcolor{black}{In order to evaluate the effectiveness of SI-ConvNets we must experiment with a dataset where objects come at variety of scales, since there is not much gain that can be obtained from learning in a scale-invariant manner when there is no scale variation in data. Unfortunately, most of the benchmark datasets for evaluating ConvNets do not fit this category. Therefore, we experiment with the modified MNIST handwritten digit classification dataset introduced in \cite{honglak} called \emph{MNIST-scale}. It consists of $28\times 28$ gray-scale images, where each digit is randomly scaled by a factor $s \in \mathcal{U}(0.3, 1)$ without making any truncation of the foreground pixels.
}

Unless otherwise noted, the architectures used in this experiment consist of two convolutional layers with 36 and 64 feature maps of 7x7 and 5x5 kernels respectively, a fully connected layer with 150 hidden nodes, and a soft-max logistic regression layer. The network architecture is modeled after the ConvNets of \cite{dropconnect} that achieve state-of-the-art on the original MNIST dataset, and we use the same pre-processing method and hyper-parameters unless otherwise noted. We don't use techniques such as data augmentation, dropout or model averaging to simplify the comparison between a convolution layer and the proposed scale-invariant convolution layer \footnote{Note that in \cite{dropconnect}, ConvNet \emph{without} dropout achieves
state-of-the-art performance along with ConvNet with dropconnect.}. Only the kernel size of the first convolution layer and weight decay parameter were re-tuned for the MNIST-Scale dataset using a subset of training data on a ConvNet and are fixed for all networks. The networks are trained for 700 epochs and the test error after 700 epochs are reported. All networks share the same random seed. The scale-invariant convolution layer uses six scales from 0.6 to 2 at a scale step of $2^{1/3}=1.26$, i.e. scales
at $1.26^{[-2:3]}$. The details of the parameters used in each experiment are provided in the supplementary materials.

\subsection{MNIST-Scale}
\textcolor{black}{In this experiment we compare our proposed network with ConvNets, the hierarchical ConvNets of Farabet \emph{et al.} \cite{farabet}, Restricted Boltzmann Machine (RBM) and its scale-invariant version of Sohn \emph{et al.} \cite{honglak}.}
\textcolor{black}{Following the experimental protocol of \cite{honglak}, we train and test each network on 10,000 and 50,000 images respectively. We evaluate our models on 6 train/test folds and report the average test error and the standard deviation. We use the same architecture and scale parameters for hierarchical ConvNets. The results are shown in Table \ref{tab:mnist-sc}. SI-ConvNet outperforms both ConvNet and hierarchical ConvNet by more than 10\%. Hierarchical ConvNet slightly underperforms ConvNet, possibly due to overfitting as it has 6 times more parameters than ConvNet/SI-ConvNet. In the scene classification context for which hierarchical ConvNet was introduced, pixel-level labels exist providing much more training data compared to the image classification settings. This result emphasizes the strength of SI-ConvNet, which achieves scale-invariant learning while keeping the number of parameters fixed.  There is large gap between RBM models and the ConvNets due to the fact that RBMs are unsupervised models and that their architecture is shallow with only one feature extraction layer. However, the relative error difference between the original and the scale-invariant version of the RBMs and ConvNets are comparable at 9.8\% and 10\% respectively for SI-RBM and SI-ConvNet. This shows that SI-ConvNet is obtaining similar improvements for being scale-invariant and that it is a good supervised counterpart of scale-invariant models. }
\begin{table}[h!]
\centering
\caption{Test Error on MNIST-Scale of various methods.}
\label{tab:mnist-sc}
\begin{tabular}{lll}
\hline
\textit{Method} & \textit{Test Error (\%) on 6 train/test fold}\\ \hline
Restricted Boltzman Machine (RBM)\cite{honglak} & 6.1 \\
Scale-invariant RBM \cite{honglak}& 5.5  \\
\textcolor{black}{Convolutional Neural Network \cite{dropconnect}} & 3.48 $\pm$ 0.23  \\
\textcolor{black}{Hierarchical ConvNets \cite{farabet}} & 3.58 $\pm$ 0.17 \\
\textcolor{black}{Scale-Invariant ConvNet (this paper)} & \textbf{3.13 $\pm$ 0.19} \\ \hline
\end{tabular}
\end{table}

\subsubsection{Measuring Invariance}
\label{sec:invar}
\begin{wrapfigure}[14]{r}{0.35\textwidth}
\vspace{-0.25\linewidth}
\centerline{\resizebox{\linewidth}{!}{\includegraphics {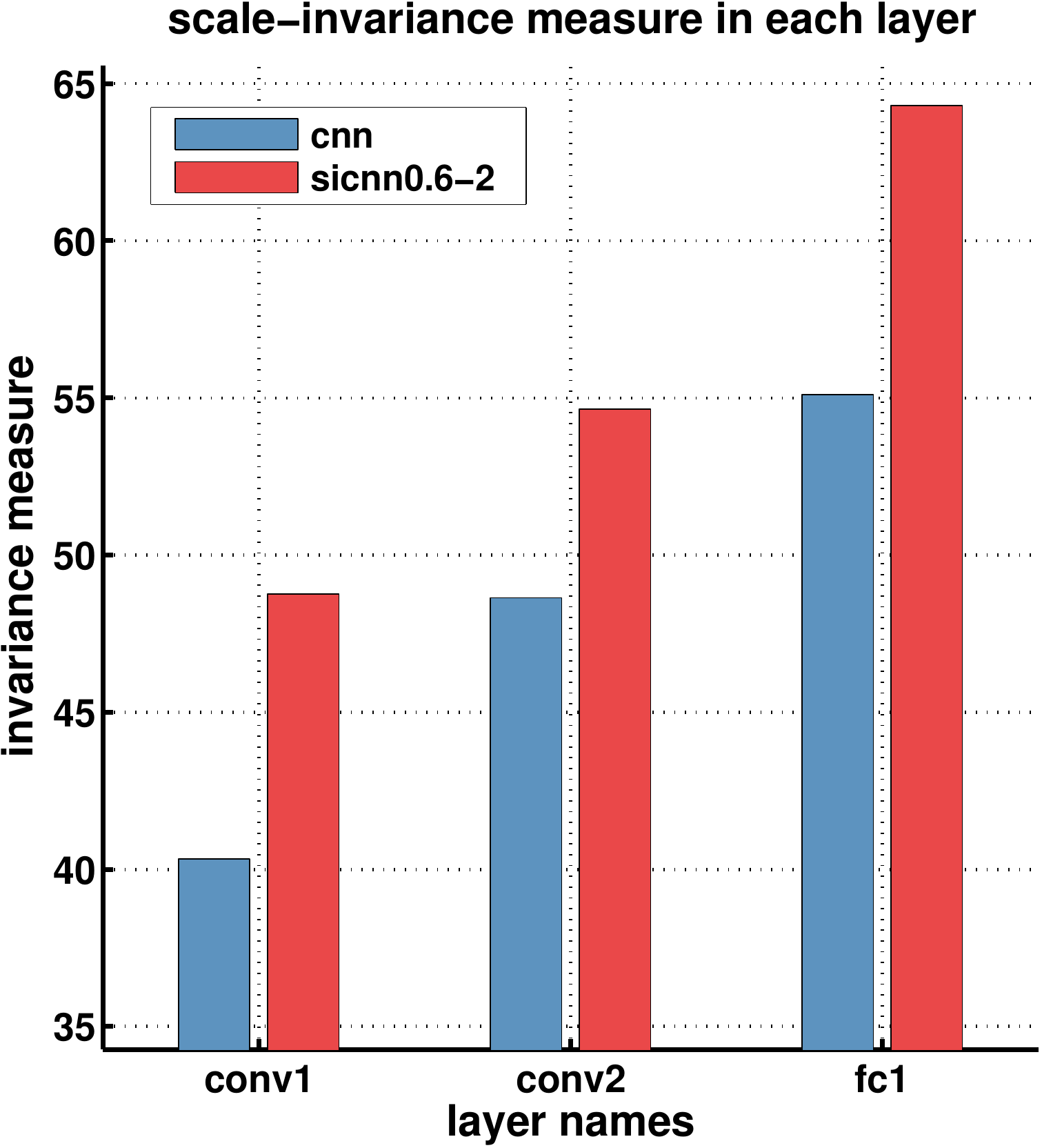}}}
\caption{\small Scale-invariance measure (larger the better).}
\label{fig:invar}
\end{wrapfigure}
We investigate the scale-invariance achieved by our model using the invariance measure proposed in Goodfellow et
al \cite{invar}.
In this method, a neuron $h_i(x)$ is said to fire when $h_i(x) > t_i$.
 Each $t_i$ is chosen so that the recall, $G(i)
= \sum|h_i(x) > t_i|/N$, over $N$ inputs is greater than 0.01.
A set of transformations $\mathcal{T}$ is applied to the images that
most activate the hidden unit, and the number of times the neuron
fires in response to the transformed inputs is recorded.
The proportion of transformed inputs that a
neuron fires to is called the \emph{local firing rate}, $L(i)$, which
measures the robustness of the neuron to $\mathcal{T}$. High $L(i)$ value
indicates invariance to $\mathcal{T}$, unless the neuron is easily
fired by arbitrary inputs. Therefore, the invariance score of a neuron is computed as the ratio of \textcolor{black}{its invariance and selectivity i.e. $L_i/G_i$}. We report the average of the top 20\% highest
scoring neurons ($p=0.2$). Please see \cite{invar} for more details.

Here $\mathcal{T}$ consists of scaling the images with values in $[0.3, 1.2]$ with step size 0.1. Figure \ref{fig:invar} shows the invariance score of ConvNet and
SI-ConvNet measured at the end of each layer. We can see that by max-pooling responses over multiple scales, SI-ConvNets produce features that
are more scale-invariant than those from ConvNets.

\subsubsection{Effect of training data and number of parameters}
\label{sec:vary-train-size}
We further evaluate SI-ConvNet by varying the number of training samples and feature maps in the first two layers. For these
experiments we report the test error on 10,000 images.

As discussed in subsection \ref{sec:fprop}, using $n$ scales in a
scale-invariant convolution layer that has $m$ kernels resembles a
network that has $nm$  kernels without actually having to
increase the number of parameters by $n$ times. One of the biggest
disadvantages of ConvNets is that it requires more training data as the number of parameters increase. By keeping the number of parameters fixed, SI-ConvNets can train a $n$ times wider and thus more powerful network at a less demanding amount of
training data. Being able to share information between the same patterns
at multiple scales further allows SI-ConvNets to learn better features
with less data. In contrast, ConvNets learn multiple filters for the
same pattern independently, and learning those filters well requires
many examples of that pattern at each scale.
\vspace{-0.2cm}
\begin{figure}[h]
\center
\subfloat[]{\resizebox{2.4in}{!}{\includegraphics{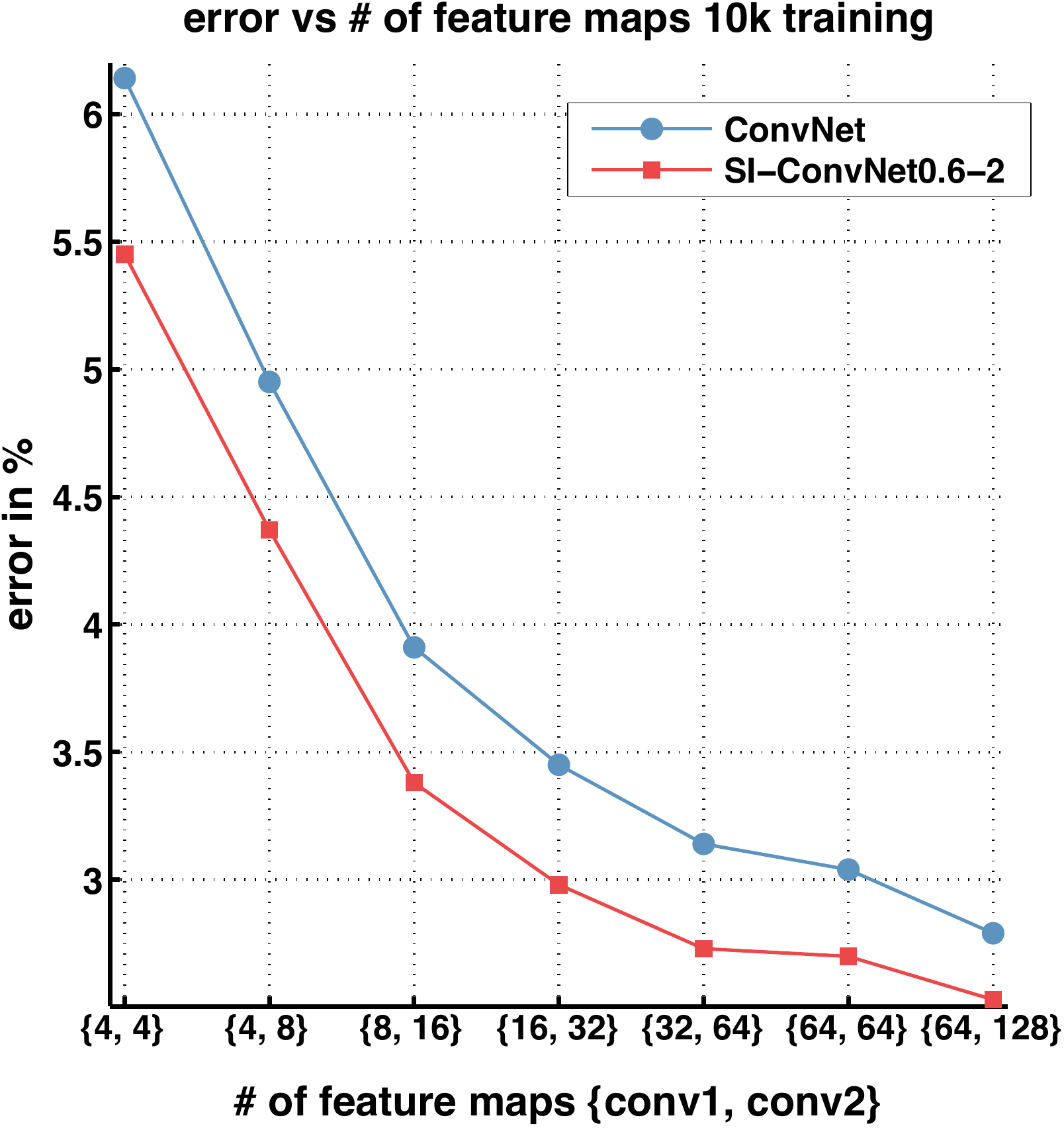}\label{fig:1a}}}
\subfloat[]{\resizebox{2.4in}{!}{\includegraphics{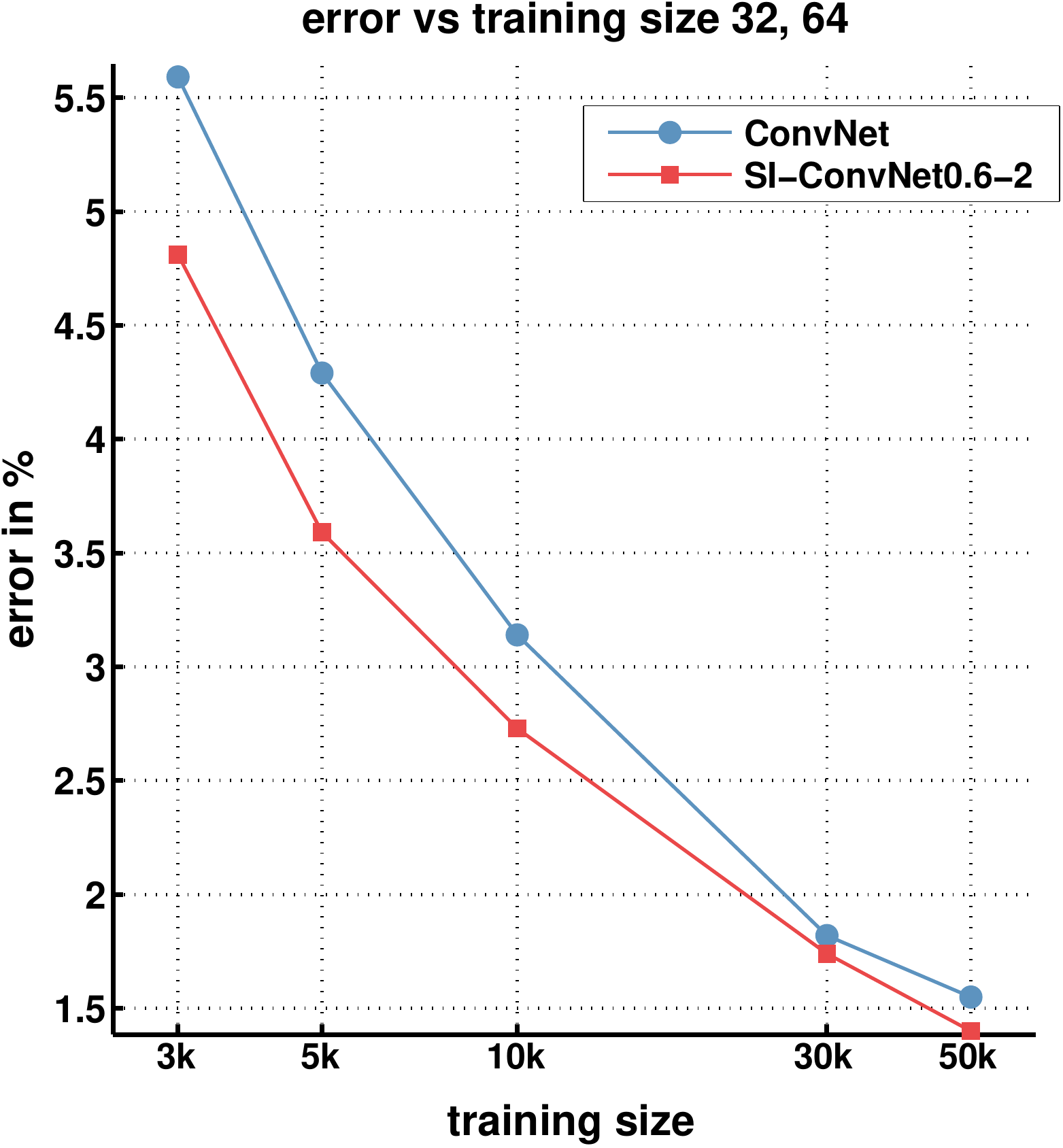}\label{fig:1b}}}
\caption{(a) Test error vs number of feature maps in the first
  two layers. (b) Test error vs amount of training data in units
of 1000, where SI-ConvNet is shown to be less prone to over-fitting.}
\label{fig:graphs}
\end{figure}

Figure \ref{fig:1a} plots test error as the number
of feature map is varied, where SI-ConvNets consistently outperforms ConvNets.
Since given enough feature maps, ConvNets can learn a feature detector
for each scale, we observe that the gap decreases as the number of feature maps
increases. Figure \ref{fig:1b} plots test error as the amount of training data changes. Again, SI-ConvNet consistently achieves
lower error than ConvNet, where their gap decreases as training data increases. This shows that SI-ConvNets can learn a better model with less training data.

\subsubsection{Robustness to Unfamiliar Scales and Scale Variation}
\label{sec:vary-scale}
In the following two experiments, we increase the image sizes of the
MNIST-scale dataset from 28x28 to 40x40 so that we can experiment with a wider range of scale
variation. In order to account for
the larger scale range, we change the scales used in SI-ConvNet from 5
scales in [0.6-2] to 5 scales in [0.5 - 2.7] for these two experiments. We train and test on 10,000 images.

\begin{figure}[h!]
\center
\subfloat[]{\resizebox{2.4in}{!}{\includegraphics{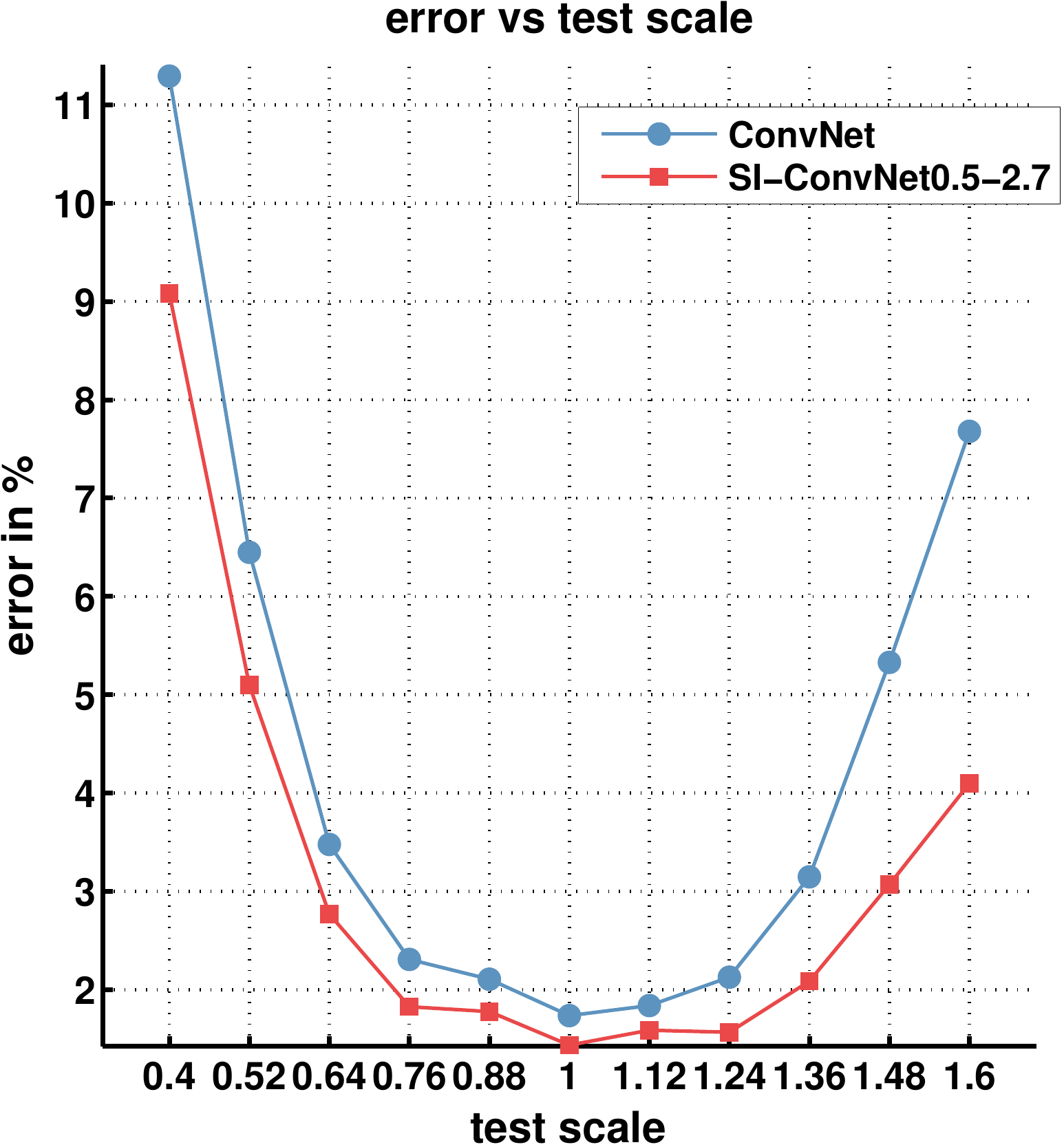}\label{fig:2a}}}
\subfloat[]{\resizebox{2.4in}{!}{\includegraphics{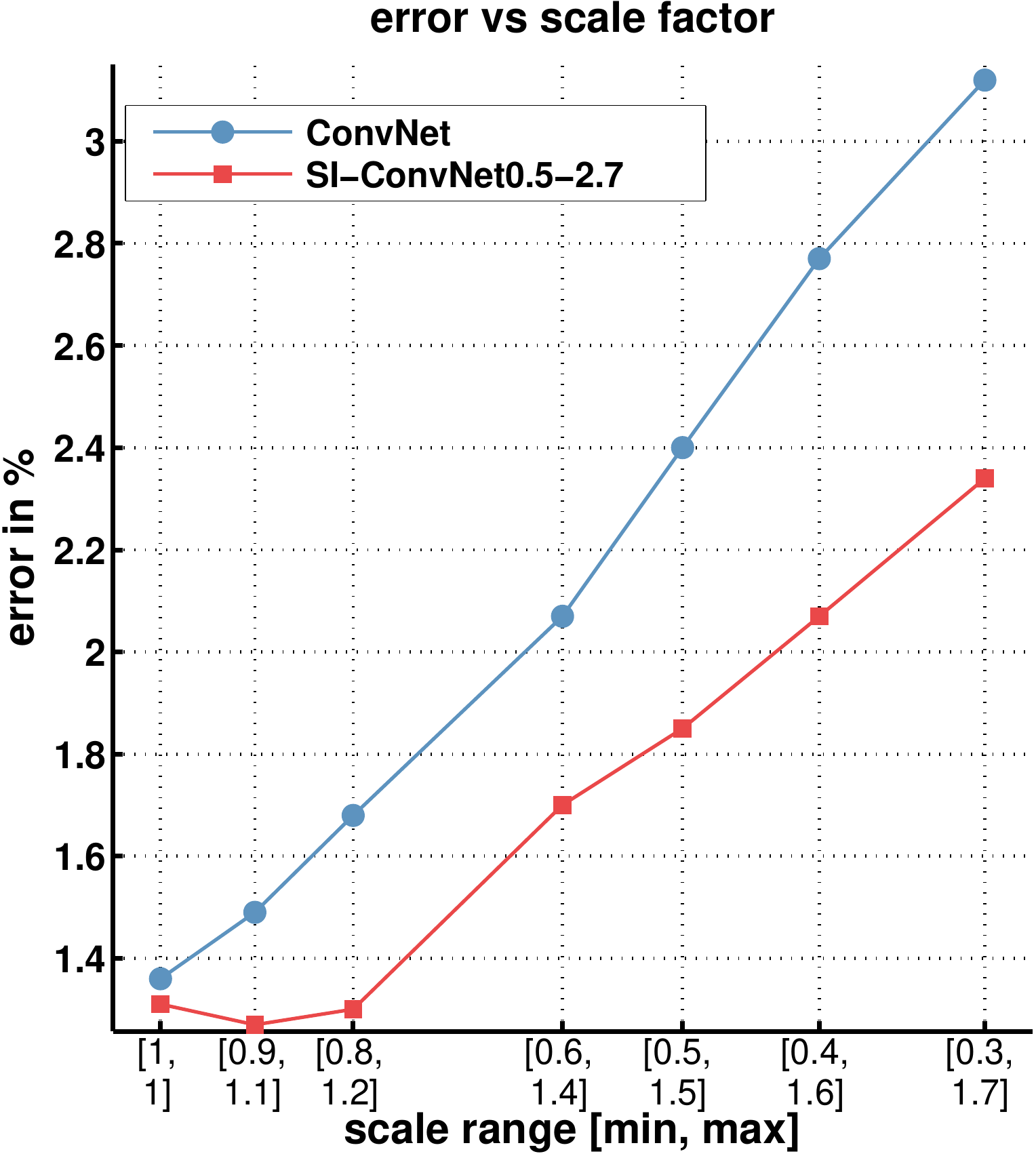}\label{fig:2b}}}
\caption{\small (a) Test error vs scale of digits at test time, where digits in the training data is scaled by a factor sampled from a Gaussian
  distribution centered at 1. The two extremes ends are about 2 standard deviation away from
  the mean. Classification is more challenging the further away the scales are from the mean, since less number of training data were available. The large gap between SI-ConvNets and ConvNets at the ends show that SI-ConvNets are more robust to images at unfamiliar scales.  (b) Test error vs range of uniform distribution used to scale training data. Lower growth of error of SI-ConvNets shows that it can learn better features given the same resources when the data complexity increases. }
\label{fig:scales}
\end{figure}

\textcolor{black}{First we evaluate the ability to correctly classify images that are less common in the training data. Here the training data is scaled by factors sampled from a Gaussian $\mathcal{N}(1, 0.24)$ rather than a Uniform distribution. The digits in the test data are scaled to one particular scale factor and we vary the test scale factor between [0.4, 1.6], which correspond to about $\pm2 \sigma$ away from the mean. The further away the test scale factor is from the mean, the more challenging the problem gets since not many training samples have been observed at that scale during the training. We expect ConvNet and SI-ConvNet to do similarly around the mean but ConvNet to get progressively worse as scale moves away from the mean as it cannot reuse the filters it learned for inputs of different scales.} As shown in Figure \ref{fig:2a}, our results verify this trend, where SI-ConvNet outperforms ConvNet even at the mean.
Average reduction in the relative error is 25\% and at the two ends of the scales 0.4 and 1.6, the relative error
reduction is 20\% and 47\% respectively. The lack of symmetry around the mean is possibly due to the fact that digit classification becomes extremely difficult even for humans when digits are very small. (For example, at the scale of 0.4, the actual digit sizes are around 8x8.)

\textcolor{black}{Next we evaluate robustness to scale variation by increasing the range of scale present in training and test data while keeping the number of parameters and training samples fixed. The scale factors are sampled from a uniform distribution in the range $[a,b]$. }
The results in Figure \ref{fig:2b} show that SI-ConvNets consistently outperform ConvNets, and that the error of SI-ConvNets increases at a
lower rate than ConvNets as the scale variation increases. This shows the weakness of ConvNets which has to learn redundant filters for digits that come at a wide variety of scales, and that SI-ConvNets is making a more efficient use of its resources in terms of the number of parameters and training data.

\vspace{-0.5cm}

\section{Conclusion}
We introduced an architecture that allows locally scale-invariant feature learning and representation in convolutional neural nets. By
sharing the weights across multiple scales and locations, a single feature detector can
capture that feature at arbitrary scales and locations. 
We achieve locally scale-invariant feature representation by pooling detector responses over multiple scales. Our architecture is
different from previous approaches in that scale-invariance is built into each convolution layer independently. Because we maintain the
same number of parameters as traditional ConvNets while incorporating the scale prior, we can learn features more efficiently with reduced chances of over-fitting. Our experiments show that SI-ConvNets outperform ConvNets in various aspects.

\vspace{-1cm}
\renewcommand{\refname}{\subsubsection*{References}}
\bibliographystyle{unsrt}
\bibliography{sicnn}
\newpage
\appendix
\section*{Appendix}
\renewcommand{\thesubsection}{\Alph{subsection}}
\subsection{Back-propagation}
\label{sec:backprop}
In order to align the notation of back-propagation to that of an
ordinary multi-layer neural network, we re-write
each step in the forward propagation of a scale-invariant convolution
layer as a matrix-vector multiplication.

Let $x^l$ be a vectorized input at layer $l$ of length $n$ ($n= hwc$
for a $h$ by $w$ by $c$ image). The spatial transformation $\mathcal{T}(x)$ can be written as a
matrix-vector multiplication of a $n$ by $m$
matrix $T$ that encodes the interpolation coefficients, where $n$ and $m$ are the dimensionality of the original and
the transformed input respectively. With bilinear interpolation, each
row of $T$ has 4 non-zero coefficients.

The convolution operation can be written as a matrix-vector
multiplication by encoding $W$ as a Toeplitz matrix. Then, the forward propagation at layer $l$ is
\begin{align}
  \label{eq:4}
  z_i^l &= \hat T_i [toep(W) (T_ix^{l-1}) + b^l]\\
  h^l &= \sigma \left (\max_{i\in \{1, \dots, n\}} \, [z_i]\right ),
\end{align}
where $T_i$ is the matrix encoding the $i$-th transformation where
images are scaled to different sizes, $\hat T_i$ is the matrix
encoding the $i$-th inverse transformation used to align the responses of
convolution on each scale. $toep(W)$  is the kernel matrix encoded as
a Toeplitz matrix.

Then, the error signal from the previous layer $\delta ^{l+1}$ can be
propagated by equations
\begin{align}
  \label{eq:5}
  \delta^{l+1}_i &= \mathbbm{1}\left ( \text{argmax}_{j \in \{1,\dots,
      n\}}    [z_j] = i\right ) \odot \sigma'(z)\\
\delta^l_i &= toep(W)^\intercal ({\hat T_i}^\intercal
\delta^{l+1}_i)\\
\delta^l &= \sum_i T_i^\intercal\delta^l_i,
\end{align}
where $\odot$ is the element-wise multiplication. Equation (3)  applies the derivative of the activation function to the
error signal and distributes the error into $n$ separate errors using
the argmax indices to un-pool the max-pooling stage. Equation (4)
propagates the error based on the linear weights $\hat T_i$ and $W$
similar to the way the error is propagated in a traditional
ConvNet. Then in Equation (5), the error is propagated through the
initial spatial transformation and is accumulated to complete the
propagation of this layer.

\subsection{Time Analysis}
\label{sec:time-analysis}
We discuss the increase in the number of convolution operations
required in a scale-invariant convolution layer compared to a
traditional convolution layer.

Given an $n \times n \times m$ input image and a kernel of size $k
\times k \times m$, a traditional convolution
layer computes the linear combination of the kernel and a local region
$(n - k + 1)^2 = \mathcal{O}(n^2)$ times (using
``valid'' convolution at the borders). For a scale-invariant convolution layer
that uses $t$ scales at a step size of $s > 1$ whose largest scale
factor is $s^k$, the input image is
scaled to $t$ different images of size $[s^kn , \cdots, s^{k-t}n]$.

So the number of linear combinations to be computed on all of the scaled
inputs is
\begin{equation}
  \label{eq:1}
  (s^kn - k + 1)^2  +
  \cdots + (n-k+1)^2 + \cdots + (s^{k-t}n - k + 1)^2.\\
\end{equation}

The summation of the quadratic terms is a geometric series,
\begin{align}
  \label{eq:2}
  &  s^{2k}n^2 + \cdots +  n^2 + \cdots + s^{2(k-t)}n^2\\
    &= n^2 s^{2k}[ 1 + \frac{1}{s^2} + \frac{1}{s^4} + \cdots +
    \frac{1}{s^{2t}}]\\
    &= n^2 s^{k} \sum_{i=0}^{t} \left(\frac{1}{s}\right)^{2i}.
\end{align}
Since $r = \frac{1}{s^2} < 0$, the series sums to
\begin{equation}
  \label{eq:3}
n^2 s^{2k} \frac{\: 1-r^{t}}{1-r}.
\end{equation}
Therefore, the number of linear combinations computed in a single scale-invariant
convolution layer is $\mathcal{O}(s^{2k} \frac{\: 1-r^{t}}{1-r}n^2)$.

For example, using the values $s = 1.26$, $t = 5$ and $k=2$ that are used in our
experiments, the series sums to 10.

\subsection{Experimental Details}
\label{sec:experimental-details}
Here we list the details of networks used in the experiments.

All inputs are pre-processed by subtracting the training mean and
the pixel values are scaled to the [0, 1] range.

The base setup is a three layer network. All experiments have this
architecture unless specified otherwise. The first layer is a (SI-)convolution layer of $7 \times 7$ kernel at
stride of one with 36 feature maps with ReLu activation function, followed by $2 \times 2$
max-pooling of stride two. The second layer is another (SI-)convolution layer of $5
\times 5$ kernel with 64 feature maps with ReLu and $3\times 3$ max-pooling of stride three.
The third layer is a fully connected layer with 150 hidden variables
with ReLu and this final output is sent to the logistic regression
layer of size 10.
The network is optimized by stochastic gradient descent of mini-batch size 128 with a fixed learning
rate of 0.01. Momentum of 0.9 and weight decay of 0.0001 are used as
regularization. Networks are trained for 700 epochs.

We tuned the kernel size of the first layer and the weight decay on a 10k validation set. After the parameters are set,
all training data were used to obtain the final network. For the experiments that are ran on $40\times 40$ images, we found the best kernel size of the first layer to be $9\times 9$. The configuration files that contains hyper-parameters and architectures
for each experiment will be available with the source code.

\end{document}